\documentclass[sigconf]{acmart}

\usepackage{multirow}
\usepackage{booktabs} 
\usepackage{ dsfont }
\usepackage{hyperref}
\usepackage{caption} 
\usepackage{cleveref} 

\usepackage[labelformat=simple]{subcaption} 

\AtBeginDocument{%
  \providecommand\BibTeX{{%
    \normalfont B\kern-0.5em{\scshape i\kern-0.25em b}\kern-0.8em\TeX}}}

\copyrightyear{2021}
\acmYear{2021}
\setcopyright{acmcopyright}\acmConference[ICMI '21]{Proceedings of the 2021 International Conference on Multimodal Interaction}{October 18--22, 2021}{Montréal, QC, Canada}
\acmBooktitle{Proceedings of the 2021 International Conference on Multimodal Interaction (ICMI '21), October 18--22, 2021, Montréal, QC, Canada}
\acmPrice{15.00}
\acmDOI{10.1145/3462244.3479923}
\acmISBN{978-1-4503-8481-0/21/10}

\begin{document}

\title{Learning Oculomotor Behaviors from Scanpath}

\author{Beibin Li}
\email{beibin@cs.washington.edu}
\affiliation{%
  \institution{University of Washington}
  \city{Seattle}
  \state{WA}
  \country{USA}
  \postcode{98195}
}

\author{Nicholas Nuechterlein}
\affiliation{%
  \institution{University of Washington}
  \city{Seattle}
  \state{WA}
  \country{USA}
  \postcode{98195}
}

\author{Erin Barney}
\affiliation{%
  \institution{Seattle Children's Research Institute}
  \city{Seattle}
  \state{WA}
  \country{USA}
}

\author{Claire Foster}
\affiliation{%
  \institution{Binghamton University}
  \city{Binghamton}
  \state{NY}
  \country{USA}
}

\author{Minah Kim}
\affiliation{%
  \institution{University of Virginia}
  \city{Charlottesville}
  \state{VA}
  \country{USA}
}

\author{Monique Mahony}
\affiliation{%
  \institution{Seattle Children's Research Institute}
  \city{Seattle}
  \state{WA}
  \country{USA}
}

\author{Adham Atyabi}
\affiliation{
  \institution{University of Colorado}
  \city{Colorado Springs}
  \state{CO}
  \country{USA}
}

\author{Li Feng}
\affiliation{%
  \department{Xiangya Hospital}
  \institution{Central South University}
  \city{Changsha}
  \state{Hunan}
  \country{China}
}

\author{Quan Wang}
\affiliation{%
  \institution{Xi'an Institute of Optics and Precision Mechanics, CAS}
  \city{Xi'an}
  \state{Shaanxi}
  \country{China}
}

\author{Pamela Ventola}
\affiliation{%
  \institution{Yale University}
  \city{New Haven}
  \state{CT}
  \country{USA}
}

\author{Linda Shapiro}
\affiliation{%
  \institution{University of Washington}
  \city{Seattle}
  \state{WA}
  \country{USA}
}

\author{Frederick Shic}
\affiliation{%
  \institution{University of Washington}
  \institution{Seattle Children's Research Institute}
  \city{Seattle}
  \state{WA}
  \country{USA}
}

\renewcommand{\shortauthors}{Li, et al.}

\begin{abstract}
Identifying oculomotor behaviors relevant for eye-tracking applications is a critical but often challenging task. Aiming to automatically learn and extract knowledge from existing eye-tracking data,  we develop a novel method that creates rich representations of oculomotor scanpaths to facilitate the learning of downstream tasks. The proposed stimulus-agnostic Oculomotor Behavior Framework (OBF) model learns human oculomotor behaviors from unsupervised and semi-supervised tasks, including reconstruction, predictive coding, fixation identification, and contrastive learning tasks. The resultant pre-trained OBF model can be used in a variety of applications. Our pre-trained model outperforms baseline approaches and traditional scanpath methods in autism spectrum disorder and viewed-stimulus classification tasks. Ablation experiments further show our proposed method could achieve even better results with larger model sizes and more diverse eye-tracking training datasets, supporting the model's potential for future eye-tracking applications.
Open source code: {\color{black} \url{http://github.com/BeibinLi/OBF}}.
\end{abstract}


\maketitle

\section{Introduction}

Human oculomotor behaviors are strongly associated with internal mental states. For example, pupil sizes change with cognitive load \cite{beatty1982task}, oculomotor gaze properties change with emotional state \cite{de2008measuring}, and patterns of visual exploration associate with skills such as executive functioning \cite{thibaut2016analogical}. Because of this diversity of both measure and application, eye tracking has been used as a multi-modal technology to study human cognition in a wide range of psychological investigations including studies of human-computer interaction (HCI). Prior work has spanned from studies of drivers' distractions in transportation \cite{zhang2006identification}, to pathologists'  mental processes in medical imaging interpretation \cite{brunye2017accuracy}, to  consumers' attention towards marketing materials \cite{lee2012attention}.  Most of these eye-tracking studies are usually task-specific and their analysis and interpretation typically requires human expert knowledge. 

The scarcity of labeled data is the primary challenge for researchers seeking to employ more domain-agnostic machine learning approaches, such as deep learning, in eye-tracking studies. While unlabeled eye-tracking data is plentiful, acquiring human subject data and assigning appropriate labels to this data has a significant overhead cost. Annotating eye-tracking data with labels such as cognitive states, presence of psychiatric conditions, or purchasing behaviors requires human reporting and expertise. Moreover, only a few large public eye-tracking datasets are available for researchers, with most eye-tracking studies including fewer than 30 participants \cite{mit-saliency-benchmark}. 
To avoid the burdensome process of manually specifying relevant oculomotor features, data scientists need a more scalable and generalizable approach such that  broader and more diverse eye-tracking applications may be enabled.

To address the scarcity of annotated data in eye-tracking studies, we created a framework called the Oculomotor Behavior Framework (OBF) to automate eye-tracking scanpath analysis. The OBF includes a neural network-based encoder to encode properties of human gaze behaviors from scanpath data of arbitrary length, and thus can facilitate many eye-tracking data mining applications. The OBF leverages four pre-training tasks (pre-tasks) and four corresponding decoders to learn from unlabelled data. We pre-train OBF on these datasets and conduct proof-of-concept experiments that demonstrate OBF is robust to data label scarcity. In the future, researchers can use the pre-trained OBF for many different downstream eye-tracking applications, including cognitive analysis, stimuli prediction,  and participant classification.

\section{Background}
\label{sec:background}

In desktop-mounted eye-tracking studies, researchers usually set up stimuli on a computer screen and use an eye-tracking device to capture participants' eye movements and the focal location of their gaze. The eye-tracking scanpath is a time series signal discretely recorded as (x, y) screen coordinates of a user's gaze.

Region-of-interest (ROI), saliency, and scanpath analyses are popular ways to analyze eye-tracking data.
ROI analysis utilizes detailed knowledge of the presented stimuli; for instance, many eye-tracking studies of ASD have found that children with ASD look less at human faces\cite{chawarska2013decreased,Frazier_Strauss_Klingemier_Zetzer_Hardan_Eng_Youngstrom_2017}. 
Saliency predictions, which also depend on stimuli, try to predict where humans would look by analyzing the shown stimulus. 
On the other hand, scanpath analysis usually does not have access to the stimulus thus treating the stimulus as a ``black-box.'' 
In this study, we focus on stimulus-agnostic scanpath methods so that the OBF could be generalized to more diverse eye-tracking stimuli and datasets. 

Traditional scanpath analyses usually first apply fixation identification algorithms to reduce the scanpath time series to a series of discrete spatially- and temporally-constrained gaze fixations. Then, experts extract various features, such as fixation convex hull shape, fixation speed, fixation duration, and cohesion \cite{wang2018cohesion} from the identified fixations. These approaches are usually effective for specific applications but often not generalizable to other eye-tracking studies.

Previous studies have used Hidden Markov models \cite{pierdicca2018user}, convolutional neural networks \cite{fuhl2021fully}, recurrent neural networks (RNNs) \cite{sims2020neural}, autoencoders \cite{elbattah2019learning} and other machine learning methods to extract features from scanpaths, but most of these studies only focus on a specific machine learning task. Similarly, random forests, SVMs, MLPs, and other machine learning approaches have been used on expert-designed features. By contrast, in this work, we focus on automatic methods to learn useful features for various eye-tracking applications. 

Unsupervised learning and self-supervised learning techniques, e.g. autoencoders, have also shown promising results in recent eye-tracking studies \cite{fuhl2021fully,elbattah2019learning}.
These studies use convolutional layers for signals \cite{fuhl2021fully} or image representations \cite{elbattah2019learning} to learn local features. 
In contrast to this previous work, to further improve the quality and generalizability of autoencoders to more diverse eye-tracking applications, in this work we use the sequence-to-sequence (seq2seq) \cite{bahdanau2014neural} design to learn time-series related features. Further, we pre-train the OBF on diverse datasets and test it for two different downstream applications. The proposed method thus yields a pre-trained model that can be used for subsequent eye-tracking applications with various scanpath lengths.

As training deep models requires a lot of data and computation power, researchers now prefer to fine-tune on pre-trained models. This trend started in the field of natural language processing (NLP), where Transformer, BERT, and GPTs pre-trained on abundant internet text data with multiple tasks were shown to facilitate adaptation to different text-mining applications. Inspired by these studies, we also pre-train our OBF from several public eye-tracking datasets. Unlike the above NLP and computer vision (CV) models, the OBF and its pre-tasks use novel and distinct pre-training methodology that are specifically designed for eye-tracking signals.

\begin{figure*}[t!]
    \centering
    \includegraphics[width=0.8\textwidth]{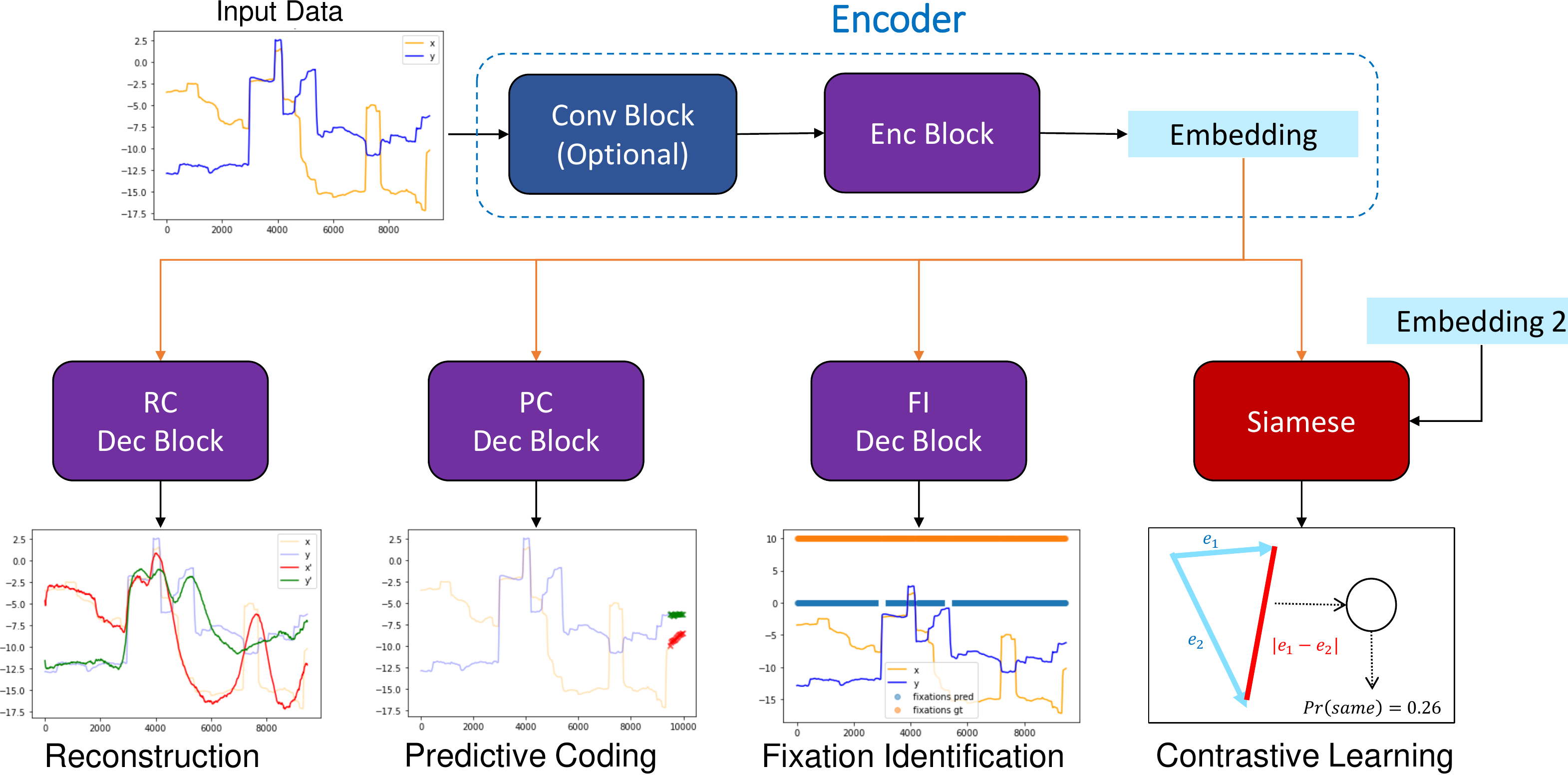}
   \caption{
The Oculomotor Behavior Framework (OBF) involves pretraining an encoder (Enc) simultaneously with multiple decoders (Dec) that are tied to different tasks. Once trained, the encoder can be quickly and efficiently retrained for adaptation to new tasks. The OBF trains all blocks simultaneously with four self-supervised learning tasks: reconstruction (RC), predictive coding (PC), fixation identification (FI), and contrastive learning (CL). Embedding 2 is calculated from another eye-tracking signal from the OBF.}
   \label{fig:obe}
\end{figure*}

\section{The OBF Pre-Training System}
\label{sec:obe}
Here, we propose the Oculomotor Behavior Framework (OBF), which includes an encoder and four decoders for four auxiliary pre-training tasks.
The OBF works as a multi-task autoencoder system, where its encoder compresses information to a fixed-length vector, and the decoders interpret the vector into different types of information.
The pre-task selection process involved eye-tracking experts’ suggestions and evaluative machine learning experiments. The four pre-training tasks are reconstruction (RC), predictive coding (PC), fixation identification (FI), and contrastive learning (CL). The RC, PC, and CL tasks use classic self-supervised learning approaches, and the FI task applies supervised learning by automatically acquiring ground truth labels from I-VT \cite{salvucci2000identifying}, a common fixation identification algorithm.

\subsection{The Encoder Network}
As shown in Figure \ref{fig:obe}, the OBF encoder contains an optional convolutional block and a sequential encoder block.
These blocks can have arbitrary sizes, dimensions, and architectures.
The sequential encoder block can be any recurrent unit (e.g., RNN \cite{rumelhart1985learning}, GRU \cite{cho2014learning}, LSTM \cite{hochreiter1997long}) or a Transformer unit \cite{vaswani2017attention}. 
If the sequential encoder block is a recurrent unit, the OBF encoder will concatenate hidden states (and also cell states for LSTM) from all recurrent layers to create the embedding.
If it is a Transformer, latent vectors from all timepoints will be used in the sequential decoders, and the last latent vector (for the $<$End-of-Sequence$>$ token) will be used as the embedding.
Modern deep learning applications have widely adopted these structures, and we refer readers to the above literature for more technical details.

\subsection{Pre-Training Tasks Overview}
The decoders for RC, PC, and FI tasks are sequential, meaning that their outputs are sequences of predictions.
These decoders have the same architecture as the sequential encoder (i.e., RNN, GRU, LSTM, Transformer) for engineering convenience.
For recurrent encoder and decoders, the forward and backward pass for each task follows the seq2seq design \cite{sutskever2014sequence}. 
For Transformer blocks, each task follows the original Transformer autoencoder design \cite{vaswani2017attention}.

Unlike the above pre-tasks, the CL task does not require sequential prediction.
The CL decoder, which deploys the Siamese network design \cite{bromley1994signature}, uses the last latent vector as input for classification or clustering purposes. 

The loss $\mathcal{L}$ for this pre-training system combines the loss 
(i.e., $\mathcal{L}_{RC}, \mathcal{L}_{PC}, \mathcal{L}_{FI}, \mathcal{L}_{CL}$) from each individual pre-tasks, where the OBF and decoders are trained simultaneously to minimize $\mathcal{L}$. The $w_{RC}, w_{PC}, w_{FI}, w_{CL}$ are weights (i.e., constants) for different pre-tasks; for simplicity, we assume they equal to $1$ for all empirical experiments.
\begin{align*}
\mathcal{L} = 
w_{RC}\mathcal{L}_{RC} + 
w_{PC}\mathcal{L}_{PC} + 
w_{FI}\mathcal{L}_{FI} + 
w_{CL}\mathcal{L}_{CL}
\end{align*}
In the following sections, we denote $\mathcal{E}$ as the encoder.
We let $\mathcal{D}$ represent the four pre-task decoders, which we individually tag with subscripts.
The input scanpath segment is $x \in \mathds{R}^{t \times 2}$ with $t$ timepoints.
For a matrix $a$, $|a|$ represents element-wise absolute value, $||a||_1$ represents the L1 norm, and $||a||_2$ represents the Frobenius Norm (L2 norm).

\subsection{Pre-Task: Reconstruction}

Input reconstruction (RC) has been widely used to train autoencoders since the Helmholtz machine \cite{hinton1995wake}.
The OBF and the RC decoder learn from recovering 5 - 10 seconds signals. 
The reconstruction loss is a standard mean squared error (MSE) between the input data and output reconstruction, which is defined as:
\begin{align*}
\mathcal{L}_{RC} =  \cfrac{1}{2 t} || \mathcal{D}_{RC}(\mathcal{E}(x)) - x ||^2_2
\end{align*}
The RC task enforces the encoder to compress important information to a low dimensional embedding, and then the RC decoder could recover the original input signal.
Encoding and decoding signals usually result in smooth signals because trivial noises only have a small impact on the reconstruction loss, as shown in Figure \ref{fig:obe} and Figure \ref{fig:results_3_tasks}. This is a desired behavior to maintain important information while neglecting trivial noises.

\subsection{Pre-Task: Predictive Coding}
Human brains tend to predict what will happen in the immediate future. Previous studies also indicate that machines could have similar predictive coding (PC) abilities \cite{huang2011predictive}, and that learning from predictions of the future can give insight into input data \cite{spratling2017review}.

Similar to the PredNet \cite{lotter2020neural} that performs next-frame prediction in video sequences, OBF and the proposed PC decoder use self-supervised learning to predict the oculomotor scanpath in the next 500 milliseconds by analyzing the last 5 - 10 seconds of the existing scanpath. During the pre-processing, we segment an eye-tracking signal into $x$ and $x'$, where $x' \in \mathds{R}^{t' \times 2}$ is the immediately following future sequence of $x$. The PC system also utilizes MSE as the loss:
\begin{align*}
\mathcal{L}_{PC} = \cfrac{1}{2 t'} || \mathcal{D}_{PC}(\mathcal{E}(x)) - x' ||^2_2
\end{align*}
Predicting the future sequence requires the encoder to learn trends and patterns from the input, which is harder than simply reconstructing the input signal.

\subsection{Pre-Task: Fixation Identification}
Previous cognitive studies have found that little or no visual information is obtained during saccades \cite{Dodge_1900}. 
Extracting information from fixations and filtering out noise from saccades are thus important steps in eye-tracking data analysis. 
Researchers have invented various fixation identification (FI) algorithms to extract fixations, a process very specific to studies of gaze and not commonly employed in other signal processing application domains.

\begin{table*}[t!]
\centering
\begin{tabular}{c|cccccc}
\toprule
Set                         & Dataset   & Eye-Tracker                & Resolution & Monitor & Scanpath Length & \# Data \\
\midrule
\multirow{3}{*}{Pre-Train}  & MSU \cite{msudataset}  & SMI iViewX Hi-Speed 1250 & 1920x1080         &         -    & 16 - 38 sec  & 2,377    \\
                            & C\&G-1 \cite{coutrot1} & Eye-Link 1000              & 1024x768          & 21 inch      & 8 - 30 sec     &  4,283 \\
                            & C\&G-2 \cite{coutrot2}  & Eye-Link 1000              & 1280x1024         & 22 inch      & 20 - 80 sec   &  592  \\
\midrule
\multirow{2}{*}{Downstream} & MIT-1003 \cite{mit1003} & ISCAN RK-464 & 1280x1024         & 19 inch      & 3 sec    &    15,045    \\
                            & Autism & Eye-Link 1000+             & 1920x1080         & 22 inch            & 15 - 23 sec   &  931 \\
\bottomrule
\end{tabular}
\caption{The datasets used in the pre-training stage and downstream application stage. The monitor resolution size, the length of each scanpath, and the number of scanpaths are listed.}
\label{tab:data}
\end{table*}

OBF approaches FI as a supervised learning task without human annotation.
The ground truth labels are approximated from I-VT (identification with velocity threshold) algorithm \cite{salvucci2000identifying}, which uses gaze velocity with expert-defined thresholds (i.e., 100 visual degrees per second, 200 ms minimum fixation length) to separate fixations and saccades. 
More accurate fixation identification algorithms exist (e.g., \cite{li2016modified,li2016optimality,hessels2017noise}), but I-VT runs in linear time, which makes it a convenient candidate for pre-training purposes.

The OBF and FI decoder perform binary classification (identify ``fixation'' or ``saccade'') for each timepoint in the scanpath sequence.
The duration of fixations in a given scanpath is usually much larger than the duration of saccades; for instance, more than 90\% of the timepoints are fixations in one of our pre-training datasets (i.e., \cite{msudataset}).
To address this class imbalance, we use weighted random sampling to ensure the number of samples from fixations and saccades are the same during the training phase.

Let us define $I_{vt}(x) \in \{0, 1\}^{t}$ as the fixation identification ground truth from the I-VT algorithm, where $0$ and $1$ stand for saccades and fixations. 
Let $m(x) \in \{0, 1\}^t$ to be a random sampling mask that can balance the number of timepoints for fixations and saccades for input sequence $x$, such that  $\sum\limits_i^t m(x)_i I_{vt}(x)_i = \sum\limits_i^t m(x)_i ( 1- I_{vt}(x)_i)$.
Then, the FI uses binary cross-entropy loss on the balanced timepoints:
\begin{align*}
 \mathcal{L}_{FI} &=  \cfrac{1}{2 \sum_i m(x)_i} \Big|\Big|\log \big(\mathcal{D}_{FI}(\mathcal{E}(x)) \big) I_{vt}(x) m(x) \Big| \Big|_1 \\
 &+  \cfrac{1}{2 \sum_i m(x)_i} \Big|\Big|\log \big(1 - \mathcal{D}_{FI}(\mathcal{E}(x)) \big) \big(1 - I_{vt}(x) \big) m(x) \Big| \Big|_1
\end{align*}
This FI task can help the OBF to learn the concept of fixations and to extract relevant features from the input scanpath sequence. This information is usually crucial for eye-tracking studies to understand human cognitive states.

\subsection{Pre-Task: Contrastive Learning}
Inspired by BERT \cite{devlin2019bert} and related studies, we design the contrastive learning (CL) decoder to decide if two scanpath segments come from the same scanpath.
In a minibatch of data, we randomly cut small segments from the original scanpath, where each segment is about 20\% - 40\% of the length of the original scanpath.
The CL decoder thus provides access to oculomotor behaviors that distinguish one scanpath from another. 

As discussed, the CL decoder uses Siamese network architecture to encode scanpath segments $x_1$ and $x_2$ to two embeddings.
Then, the CL decoder calculates the absolute distances for each dimension between the two segment embeddings, which are fed into a multi-layer perceptron (MLP) for classification.
We set the ground truth label $S(x_1, x_2)$ equal to $1$ if $x_1$ and $x_2$ are from the same scanpath; otherwise, we set $S(x_1, x_2)$ to $0$.
The CL task is to approximate the function
$\mathcal{D}_{CL}(|\mathcal{E}(x_1) - \mathcal{E}(x_2)|) = S(x_1, x_2)$, and the loss is:
\begin{align*}
\mathcal{L}_{CL} &= \log \big(\mathcal{D}_{CL}(|\mathcal{E}(x_1) - \mathcal{E}(x_2)|) \big) S(x_1, x_2) \\
&+  \log \big(1 - \mathcal{D}_{CL}(|\mathcal{E}(x_1) - \mathcal{E}(x_2)|) \big)  \big( 1 - S(x_1, x_2) \big)
\end{align*}

\subsection{Pre-Training Data}

\subsubsection{Pre-Training Dataset}
Merging datasets is a standard strategy in modern deep learning studies for expanding and diversifying training datasets. Following this strategy, we assemble a multi-source eye-tracking dataset with different types of presentation stimuli, experiment length, eye trackers, monitors, hardware setups (e.g., with/without chin rest), and participants. We pre-train OBF on three datasets (Table \ref{tab:data}):
the Lomonosov Moscow State University (MSU) study collected eye-tracking data from 48 participants, each of whom watched 41 video sequence;
the Coutrot \& Guyader (C\&G)-1 study recruited 72 participants to watch 60  stimuli in four different audio-visual conditions, where 18 participants watched each audio-visual condition;
the C\&G-2 study recruited 40 participants in 2 conditions, where 20 participants watched each condition.
We choose these datasets, because they recruited many participants from various setups, which could enrich the diversity for our pre-training stage.

To the best of our knowledge, no participants in the pre-training data appeared in the downstream application.
During batch processing in the pre-training phase, a batch only contains signals from one database so that the CL task could be more challenging for the OBF to learn.

\subsubsection{Data Representation}
To create a general method for all types of eye-tracking data, we pre-process the data by using the following criteria.
All signals are resampled to 60 Hz with bilinear interpretation because previous studies suggest many oculomotor behaviors could be analyzed within 60 Hz  \cite{leube2017sampling,kredel2017eye}.
If the gaze data is binocular, the signal from the left eye and right eye is averaged.
The coordinates of the screen center are assigned to (0, 0). Coordinates (x, y) represent horizontal and vertical locations, respectively.
The coordinate unit is normalized to visual degrees because pixels are not meaningful across different eye-tracker setups.
Missing data due to equipment failure or blinks are filled with bilinear interpolation. However, when more than 50\% data are missing in a scanpath, the whole scanpath is discarded.
Gaze points that are more than 10 visual degrees off-screen are marked as an extreme value (i.e., -180 degree); this usually indicates the participant is distracted during the experiment.

\subsection{Pre-Training Engineering Details}
Unless otherwise specified, we use the following OBF encoder, which we tuned to balance run-time and performance: 
\begin{itemize}
    \item The OBF encoder contains a convolutional layer (with kernel size 7 and 30 output channels), a leaky ReLU activation layer, a residual connection to the input, and an average pooling layer (with kernel size 2).
    \item The OBF encoder uses a 2-Layer GRU unit with 128 hidden neurons in each layer.
    \item The RC, PC, and FI decoders in the OBF have the same architecture as the OBF encoder's sequential block (i.e., 2-Layer GRU unit with 128 hidden neurons in each layer).
    \item The CL decoder has a hidden layer with 128 neurons, a sigmoid activation layer, and a batch normalization layer.
\end{itemize}

\begin{figure}[h!]
    \centering
    \includegraphics[width=0.40\textwidth]{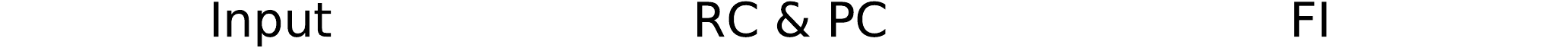}
    \includegraphics[width=0.40\textwidth]{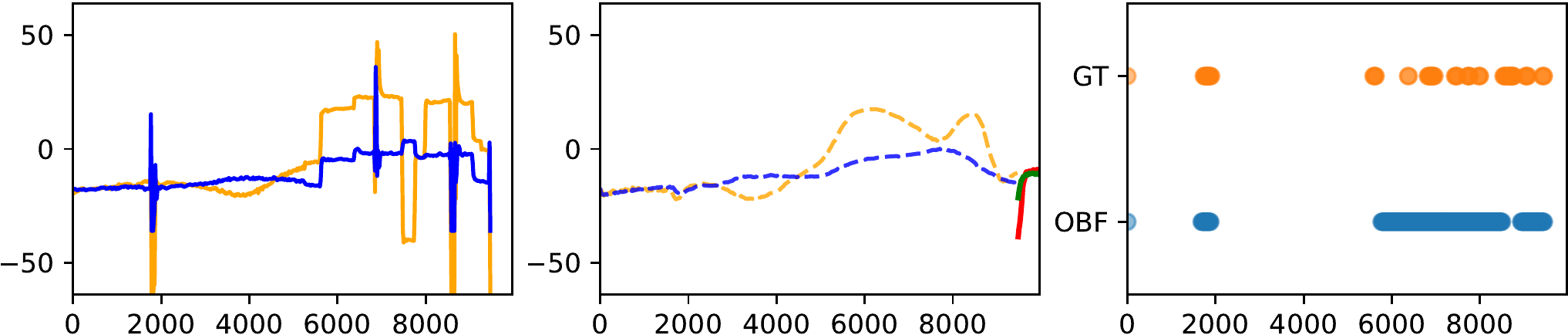}
    \includegraphics[width=0.40\textwidth]{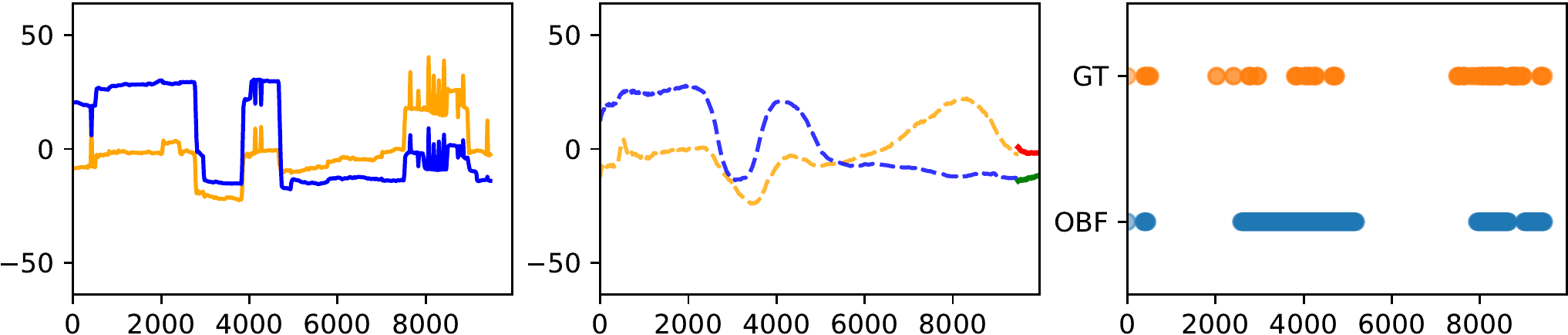}
    \includegraphics[width=0.40\textwidth]{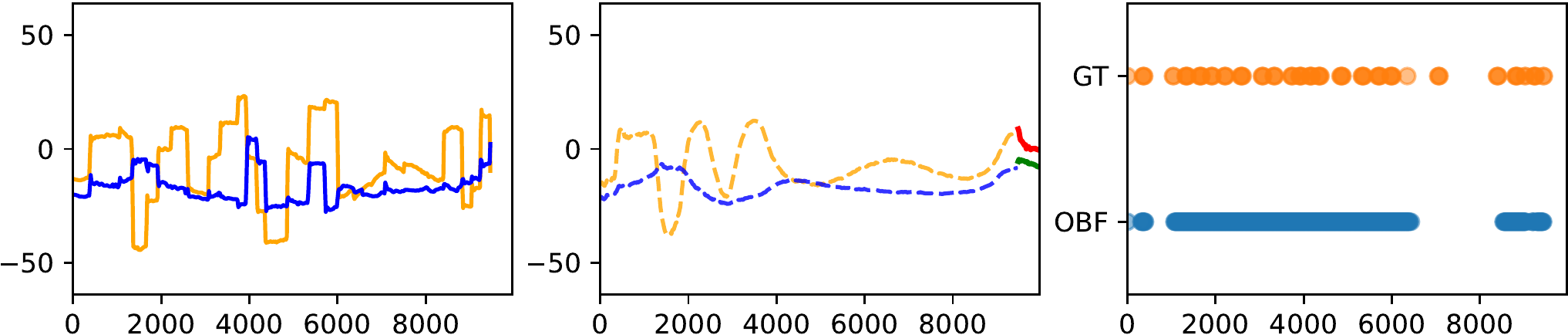}
    \caption{
Results from the Pre-Train Stage: the proposed encoder encodes the input signal (with size $\sim$ 8 KB) to an embedding vector (with size 0.5 KB). We show the input signal (x signal in orange and y signal in blue) in the first column, the reconstructed signal in the second column, and the predicted future signal in green and red lines in the second column. 
The third column shows the saccades from the fixation
identification (FI) task, where the top rows is ground truth (GT), and the bottom row is the prediction
from our model.}
    \label{fig:results_3_tasks}
\end{figure}

We pre-train the OBF for 500 epochs with a learning rate of 0.001. The learning rate is halved every 100 epochs. The gradient for each neuron is clipped to 0.5 to avoid gradient exploding. 
We set the mini-batch size to 64 based on computational efficiency considerations.

For the standard supervised learning experiment, we use 80\% of the scanpaths for training and the rest for validation. The following metrics are calculated for the validation set:
for the RC and PC tasks, we use mean Euclidean distance as the evaluation metric.
for the FI task, we use Area-Under-ROC-curve (AUC) score, because the testing labels are unbalanced.
for the CL task, we use accuracy.

Figure \ref{fig:results_3_tasks} shows an example of  inputs and outputs from the pre-training stage.
A detailed analysis of OBF is given in Section \ref{sec:ablation}. OBF's  pre-task results are shown in Table \ref{tab:model_bone} and Table \ref{tab:model_size} where they are compared to results of educated guessing (mean for regression tasks, or majority for classification tasks) and state-of-the-art methods.  We re-implement the fully convolutional variational autoencoder (C-VAE) described in \cite{fuhl2021fully} and use it with the proposed OBF pre-training tasks. OBF performs well on all four pre-tasks even with a small embedding vector.

\section{Example Applications}
\label{sec:application}

To illustrate the usefulness and potential applications for the OBF, we conduct experiments on two downstream applications - stimulus prediction and autism classification.
The pre-trained OBF achieved promising results during supervised learning and transfer learning in both downstream applications. 
We also conduct meta-learning experiments for the stimulus prediction application, where the OBF can work with more advanced learning techniques to boost classification performance.

\subsection{Stimulus Prediction}

We use the MIT-1003 dataset for this stimulus prediction experiment, where we would like to predict which stimulus the user was watching based on her/his gaze scanpaths.
The MIT-1003 dataset has 15,045 scanpaths, and was originally designed to study saliency maps of images.
The large number of scanpaths in this dataset can help us to evaluate the effects of the pre-training process in detail.
In total, 15 users participated the study, with each participant watching 1003 stimuli (i.e., images) for 3 seconds on each stimulus.
This experiment illustrates how pre-training and fine-tuning the OBF could improve downstream application performance when the number of users in an eye-tracking study is low.

We use a multi-layer perceptron (MLP) for the standard supervised learning experiment.
The input signal first feeds into the OBF (corresponding to the dashed blue box in Figure \ref{fig:obe}); then, the MLP predicts which stimulus the user was watching based on the embedding calculated by the OBF. 
The MLP contains 2 hidden layers (with 256 and 512 neurons each): for each hidden layer, we add a dropout layer (with 0.5 probability), sigmoid activation, and batch normalization.

We also performed a metric-based meta-learning experiment with prototypical network \cite{snell2017prototypical} for this downstream application. We reserved 200 stimuli for meta-training because most eye-tracking studies involve fewer than 200 stimuli (as shown in \cite{mit-saliency-benchmark}). We train all models for 100 epochs with 100 iterations per epoch. The model used in this experiment has the same structure as the one in the supervised learning experiment, and we use 128 neurons as the embedding space for the metric-based meta-learning.

For both the supervised and metric based experiments, we compare the OBF model with the C-VAE model (with the same structure as proposed in \cite{fuhl2021fully}), and a baseline method that is not pre-trained with the four decoders as described in Section \ref{sec:obe}.

To better understand the effect of pre-training, we perform $c$-way $k$-shot classification experiments to learn from $c$ stimuli and $k$ users, where $c \in [10, 100, 1003]$ and $k\in [1, 3, 5, 10]$.

The OBF pre-trained model performs  1.1 - 2 times better than the baseline in all the settings.
This result suggests that the pre-trained OBF may be of benefit to the training of future eye-tracking applications, despite those applications being developed from data from only a few participants.

\begin{table}[h!]
\centering

\begin{subtable}[t]{0.45\textwidth}
\begin{tabular}{c|c|cccc}
\toprule
Experiment                               & Method       & 1-shot        & 3-shot        & 5-shot        & 10-shot       \\
\midrule
\multirow{3}{*}{Supervised}              & C-VAE \cite{fuhl2021fully} & 0.23          & 0.18          & 0.19          & 0.34          \\
                                         & No Pre-Train  & 0.21          & 0.22          & 0.28          & 0.30          \\
                                         & Ours (OBF)   & \textbf{0.25} & \textbf{0.28} & \textbf{0.41} & \textbf{0.63} \\
\midrule
\multirow{3}{*}{Metric-based} & C-VAE \cite{fuhl2021fully} & 0.40          & 0.60         & 0.65          & 0.71          \\
                                         & No Pre-Train  & 0.50          & 0.65          & 0.72          & 0.77          \\
                                         & Ours (OBF)   & \textbf{0.52} & \textbf{0.70} & \textbf{0.73} & \textbf{0.79} \\
\bottomrule
\end{tabular}
\caption{10-way Classification}
\end{subtable}

\begin{subtable}[t]{0.45\textwidth}
\begin{tabular}{c|c|cccc}
\toprule
Experiment                               & Method       & 1-shot        & 3-shot        & 5-shot        & 10-shot       \\
\midrule
\multirow{3}{*}{Supervised}              & C-VAE \cite{fuhl2021fully} & 0.04          & 0.11          & 0.15          & 0.24          \\
                                         & No Pre-Train  & 0.04          & 0.08          & 0.14          & 0.36          \\
                                         & Ours (OBF)   & \textbf{0.06} & \textbf{0.14} & \textbf{0.32} & \textbf{0.44} \\
\midrule
\multirow{3}{*}{Metric-based} & C-VAE \cite{fuhl2021fully} & 0.16             &      0.36        &       0.41        &      0.47        \\
                                         & No Pre-Train  & 0.12          & 0.28 & 0.31 & 0.39 \\
                                         & Ours (OBF)   & \textbf{0.18} & \textbf{0.37} & \textbf{0.44} & \textbf{0.51} \\
\bottomrule
\end{tabular}
\caption{100-way Classification}
\end{subtable}

\begin{subtable}[t]{0.45\textwidth}
\begin{tabular}{c|c|cccc}
\toprule
Experiment                               & Method       & 1-shot        & 3-shot        & 5-shot        & 10-shot       \\
\midrule
\multirow{3}{*}{Supervised} & C-VAE \cite{fuhl2021fully} & 0.01          & 0.04             & 0.08          & 0.14          \\
                            & No Pre-Train  & 0.01          & 0.05          & 0.15          & 0.30          \\
                            & Ours (OBF)   & \textbf{0.02} & \textbf{0.08} & \textbf{0.17} & \textbf{0.31} \\
\bottomrule
\end{tabular}
\caption{1003-way Classification with Supervised Learning.
Metric-based meta-learning is not available for this experiment, because the meta-testing set would be empty.}
\end{subtable}

\caption{ Stimulus Prediction Experiment Testing Results for the MIT-1003 Experiment:
Learning from k-shot examples for each stimulus. 
We performed standard supervised learning and metric-based meta-learning for different scenarios.
}
\label{tab:mit_rst}
\end{table}

\subsection{Autism Classification}
To further illustrate the usefulness of OBF's learned representations, we apply OBF to a challenging real-world eye-tracking dataset acquired from children with and without autism spectrum disorder (ASD).
Researchers can extract embeddings from the OBF and then apply traditional machine learning approaches directly without fine-tuning the OBF.

In total, 49 children from this study were included in the examined dataset, with thirty-eight of the children diagnosed with ASD and the rest typically developing (TD) children. 
These children watched 19 stimuli on a 22-inch screen monitor in a room together with their parents and an experimenter. The monitor-to-head distance is about 65 centimeters, and the whole experiment takes about 11 minutes.
The data acquisition protocol was approved by the medical ethics Institutional Review Boards from the Yale University and from the Seattle Children's Research Institute. Eye-tracking data was collected under parents' and participants' consent and/or assent, as appropriate. Handling of data was compliant with Health Insurance Portability and Accountability Act (HIPAA) and institutional data security guidelines.
The experiment protocol follows Good Clinical Research Practice (GCP).

We would like to classify the participants into two groups (ASD v.s. TD) by analyzing their scanpaths. For a given participant's data, the OBF first extracts the embedding for each scanpath and then concatenates all the embeddings together to form a final long embedding. 
We use the lasso for the classification task because the lasso's L1 regularization encourages the model to ignore redundant features from the long embedding vector.

We compare the features extracted from the pre-trained OBF with expert-guided features.  
The expert-guided scanpath features include the number of fixations, total fixation duration, saccade speeds, and average fixation speed, all of which have been used in previous autism research (e.g., discussed in \cite{sasson2010}).
We also compare our results with SGIN \cite{li2020selection}, which employs deep neural networks on features extracted from traditional ROI and scanpath analyses, to provide state-of-the-art performance benchmarks in the autism classification task. 

We use 5-fold cross-validation to evaluate these features, as shown in Table \ref{tab:asd}.
The representations learned by OBF outperform the expert-guided features by a large margin. 
 The OBF shows promising results even though it employs fewer stimuli and participants compared to \cite{li2020selection}. 
More detailed analyses on how the pre-tasks could influence OBF's performance on autism classification are shown in Table \ref{tab:pretask}.

\begin{table}[h]
\centering
\begin{tabular}{l|ccc}
\toprule
Method                & Accuracy & AUC & F-1 \\
\midrule
Expert Features from \cite{sasson2010}   & 0.74 & 0.68 & 0.82  \\
SGIN (with more data) \cite{li2020selection}                  & 0.78 & \textbf{0.83} &  0.83  \\
Ours (OBF)                   & \textbf{0.80} & \textbf{0.83} &  \textbf{0.88}  \\
\bottomrule
\end{tabular}
\caption{Results for Autism Classification}
\label{tab:asd}
\end{table}

\section{Discussions and Ablations}
\label{sec:ablation}
Here, we perform ablation studies on different backbone units, model size, pre-tasks, and pre-training datasets. 
For the following analysis, we evaluate the inference run-time on an Nvidia GTX 1080 for a 600-length (10-second) scanpaths.
We evaluated 1450 scanpaths without batch processing and reported the average run-time in milliseconds.
Future studies can trade-off the run-time and performance for their specific needs.

\subsection{Effects from OBF Backbone}

We validate the RNN, the GRU, the LSTM, and the Transformer for the sequential encoder and decoder blocks for the OBF (Table \ref{tab:model_bone}). 

\begin{table}[h]
\centering
\begin{tabular}{l|rr|cccc}
\toprule
Enc Block & Para & T & \begin{tabular}[c]{@{}c@{}}RC \\ Dist\end{tabular} & \begin{tabular}[c]{@{}c@{}}PC\\ Dist\end{tabular} & \begin{tabular}[c]{@{}c@{}}FI \\ AUC\end{tabular} & \begin{tabular}[c]{@{}c@{}}CL\\ ACC\end{tabular} \\
\midrule
GRU           &   163k    &  4.62  &  \textbf{4.06}   &  \textbf{6.38}  &   \textbf{0.80}     &  \textbf{0.85}       \\
RNN           &   55k     &  4.61  &  12.0   &  7.68  &   0.52     &  0.63       \\
LSTM          &   217k    &  4.71  &  4.55   &  6.45  &   0.72     &  0.83       \\
Transformer   &   343k    &  2.59  &  8.93   &  7.12  &   0.61     &  0.76       \\
\midrule                                                           
GRU (no Conv) &   150k    &  7.68  &  4.35   &  \textbf{6.38}  &   \textbf{0.80}    &   0.84      \\
\midrule
Educated Guess &   &    &  11.8   &  11.8  &   0.5   &   0.5      \\
C-VAE \cite{fuhl2021fully} with OBF &  \textbf{15k} &  \textbf{1.08} & 8.18   & 6.54  &   0.65   &   0.79      \\
\bottomrule
\end{tabular}
\caption{Effects from different model backbones, where we use a 2-layer unit with 128 hidden neurons in each layer for these models.
The number of trainable parameters (para) in the encoder and the inference run-time (T) in milliseconds are reported.
The Euclidean distance (Dist), AUC score, and accuracy (ACC) are evaluated for the pre-tasks.
}
\label{tab:model_bone}
\end{table}

The GRU and the LSTM perform well on all pre-tasks, and the GRU is more memory-efficient than the LSTM. 
Standard RNN units perform the worst among all tested backbones. 
In our pre-train experiments, the input sequence can contain over 600 timepoints (10 seconds of data), which might cause the vanishing of gradient problem inside the standard RNN unit. The memory gate inside GRU and LSTM can help mitigate this problem.

Surprisingly, the Transformer model performs worse than recurrent methods in the pre-tasks. We evaluated the Transformer again with the MIT-1003 downstream application, and its performance is similar to GRU OBF's performance.
The positional encoding might smooth useful signals when the hidden dimension is not high enough, while the RC, PC, and FI pre-tasks rely heavily on the position information.

While the OBF model performs similarly on the PC and FI tasks regardless of whether the optional convolution block is included, adding the convolution block is still advantageous because it makes the recurrent block more efficient. The convolution block halves the number of time-steps required and thus significantly reduces OBF's run-time. 

While the C-VAE \cite{fuhl2021fully} has the fastest inference runtime, its performance is slightly lower than the GRU and the LSTM units even after we control for model size (Table \ref{tab:model_size}, first row). Additionally, the C-VAE could not encode various-length scanpaths into fixed-length embeddings, which is a potential limitation given that scanpaths can have radically different lengths. 
Nevertheless, researchers could use the fully-convolutional design for application with similar-length scanpaths.

\subsection{Effects from OBF Model Size}

\begin{table}[h]
\centering
\begin{tabular}{l|rr|cccc}
\toprule
Enc Block & Para & T & \begin{tabular}[c]{@{}c@{}}RC \\ Dist\end{tabular} & \begin{tabular}[c]{@{}c@{}}PC\\ Dist\end{tabular} & \begin{tabular}[c]{@{}c@{}}FI \\ AUC\end{tabular} & \begin{tabular}[c]{@{}c@{}}CL\\ ACC\end{tabular} \\
\midrule
2 x 32  &   15k        &  4.65   &  5.46   &  6.50  &   0.71   &    0.81     \\
2 x 64  &   46k        &  4.67   &  4.52   &  6.53  &   0.71  &    0.81  \\
2 x 128 &   163k       &  4.62   &  4.06   &  6.38  &   0.80     &  \textbf{0.85}       \\
2 x 256 &   620k       &  4.72   &  \textbf{3.77}   &  \textbf{6.28}  &  \textbf{0.82}   &   \textbf{0.85}      \\
\midrule
1 x 128 &   64k        &  2.50   &  4.89   &  6.39  &   0.75  &    0.83     \\
2 x 128 &   163k       &  4.62   &  4.06   &  6.38  &   0.80     &  0.85       \\
3 x 128 &   262k       &  7.02   &  \textbf{3.52}           &  \textbf{6.30}       &  0.84    & \textbf{0.86}       \\
4 x 128 &   361k &  11.89   &  3.67           &  6.42       &  \textbf{0.89}    & 0.85       \\
\bottomrule
\end{tabular}
\caption{Effects from different model size. 
A GRU unit with a convolutional layer are used across these experiments.
The first column represents (the number of layers) x (the number of neurons in each layer) for the GRU unit.
The number of trainable parameters (param) and inference run-time (T) in milliseconds are also reported. 
}
\label{tab:model_size}
\end{table}

As shown in the above table, small model sizes (e.g., 2 layers x 32 hidden units) can also yield satisfiable performance.
Thus, the OBF has potential for future mobile-efficient real-time inference on cellphones and portable devices.
Still, larger models can perform better than relatively smaller models.

Increasing the number of layers has a significant impact on runtime, while increasing the number of neurons has a significant impact on model size. 
With a two-layer GRU, the number of neurons in each hidden layer has very limited effects on the inference run-time.
The marginal improvement from model size diminishes with more than 4 layers or 256 neurons, which could be a limitation from the pre-train dataset size.

\subsection{Effects from Pre-Tasks}
\label{sec:pre_task_ablation}

To evaluate how each pre-task could influence OBF's performance in downstream applications, 
we conduct the following ablation experiments by removing one pre-task in each experiment. 

\begin{table}[h]
\centering
\begin{tabular}{c|ccc|ccc}
\toprule
\multirow{2}{*}{} & \multicolumn{3}{c|}{10-Shot Accuracy}          & \multicolumn{3}{c}{Autism-Classification}    \\
                  & 10-w        & 100-w       & 1003-w      & Acc          & F-1            & AUC           \\
\midrule
All Tasks         & \textbf{0.65} & \textbf{0.48} & \textbf{0.31} & \textbf{0.80} & \textbf{0.88} & \textbf{0.83} \\
No-RC             & 0.4           & 0.43          & 0.29          & \textbf{0.80} & 0.87          & 0.71          \\
No-PC             & 0.56          & 0.47          & \textbf{0.31} & 0.76         & 0.85          & 0.83          \\
No-FI             & 0.58          & 0.45          & 0.30          & 0.71         & 0.82          & 0.7           \\
No-CL             & 0.33          & 0.37          & 0.26          & 0.76         & 0.84          & 0.80          \\
\bottomrule
\end{tabular}
\caption{Ablation for pre-train tasks. 
For the 10-shot supervised learning experiment with MIT-1003 dataset, we evaluate the performance in 10-, 100-, 1003-way classification tasks.
For the Autism classification experiment, we evaluate the accuracy, F-1, and AUC scores.
}
\label{tab:pretask}
\end{table}

Results from Table \ref{tab:pretask} indicate that all four of these pre-tasks positively contribute to OBF's downstream performance. For the MIT-1003 stimulus prediction application, the CL and RC tasks contribute the most, while the FI task contributes the least.
This phenomenon  might be caused by the fact that fewer fixations are presented in the short period (3 seconds) of time, but the locations (predicted in the reconstruction task) and similarities (implicitly learned in the contrastive learning task) are more valuable for this stimulus prediction task. 
On the other hand, the FI task contributes the most to the autism classification application, which might be caused by the fact that children with and without ASD have distinct fixation behavior and cognitive loads \cite{shic2008amorphous}.

While all four of the pre-tasks are helpful for future downstream applications, 
eye-tracking and machine learning researchers could design other relevant pre-tasks to improve OBF's performance. We also examined the following pre-training tasks: “data source prediction”, “prediction of number of fixations/saccades”, and many others. We found that the data source prediction task was too easy because participants engaged in very different tasks for different data sets (e.g., visual search, video watching, etc.), and these tasks were highly distinguishable based on trivial features (e.g. average gaze position). Number of fixations and saccades are highly correlated with the length of the signal, and in our testing the OBF model failed to develop generalized representations of fixations and saccades based on fixation/saccade count prediction. We replaced these tasks with an explicit fixation identification task, which allowed for much richer physiologically-meaningful representations.

\subsection{Effects from Diverse Datasets}
\label{sec:diverse_datasets}
Adding more diverse datasets in the pre-train stage could result in better performance in downstream applications (details in Appendix). 
As giant billion-parameter pre-trained models have been released for computer vision and natural language processing in recent years, larger pre-trained models for eye-tracking data could be possible in the future when more public datasets are fed into the OBF.

\subsection{Broader Impact}
The OBF has promising potential to aid human experts in cognitive analysis, autism classification, healthcare applications, and human-computer interactions with eye-tracking technology.
The OBF method could also compress eye-tracking data into a lower-dimension embedding, which can help database systems to store valuable data more efficiently. 

Our code and pre-trained models are open-sourced in \url{http://github.com/BeibinLi/OBF}. 
Moreover, as most deep learning packages already include convolutional,
recurrent, and Transformer layers, re-implementing OBF only requires a minimal amount of coding effort. This simplicity reduces obstacles for future applications so that researchers from various backgrounds could easily understand and apply the OBF.

However, the OBF could have a malignant impact on privacy if it is used without users' permission or acknowledgment.
In the future, privacy-preserving deep learning techniques, such as federated learning, could help alleviate OBF's privacy vulnerability. 
We advocate responsible and human-centered deployments of these deep learning technologies. To the best of our knowledge, all eye-tracking data used in this study are agreed upon by the participants, and all these data are anonymized. 

\subsection{Limitations and Future Directions}

In this study, we used datasets collected from desktop-mounted eye-trackers because most physiology studies use similar eye-trackers. However, we believe our approach would be easily applied to head-mounted eye-trackers (e.g., eye-trackers in Google Glass, Oculus). 

Focusing on the pre-training methodologies, we show that OBF can perform well even with traditional supervised learning, transfer learning, and meta-learning in downstream applications. In the future, more learning techniques can be tested to further improve the performance in these downstream applications.

\section{Related Studies}

Only a few studies focus on creating a generic deep learning method for different eye-tracking paradigms. Crossed Eyes \cite{10.1145/3448017.3457386} studied domain adaptation across five different stimulus paradigms by using hand-crafted features. Our pre-trained model, which can extract more sophisticated features, has the potential to work concurrently with their proposed domain adaptation method.
Two of our pre-training tasks are similar to the segmentation and the reconstruction tasks proposed by Fuhl et al. \cite{fuhl2021fully}. However, our pre-trained model is more flexible and generalizable, having the ability to handle scanpaths with different lengths, to learn multiple tasks simultaneously in pre-training, and to solve a wide array of downstream eye-tracking problems.

Some recent studies \cite{stober2015deep,zhang2021distilling} applied pre-training strategies on electroencephalogram (EEG) signals, which share some data property similarities to eye-tracking signals. In this study, we proposed a pre-training framework with a deep learning model that can be generalized to different eye-tracking applications. 
Future studies can utilize the OBF to pre-train a single meta model for different bio-signals and multi-modal interaction signals (including eye-tracking, affective computing, user mouse movements, EEG, heart rate, functional magnetic resonance imaging, etc.), exploiting common characteristics shared across human-generated and human-derived data signals.

\section{Conclusion}

To address the common data scarcity problem and to alleviate the analytical burden in eye-tracking studies, we create an automatic deep learning-based feature extractor to analyze oculomotor scanpath signals. Our proposed OBF utilizes the state-of-the-art methods in recent big data advancements, including convolutional, recurrent, and Transformer neural networks. Our novel pre-training methodology combines both unsupervised and supervised approaches so that the OBF can learn to encode important oculomotor behaviors from unlabeled scanpath data. The OBF can predict which stimulus a user was watching, and it is 1.1 to 2 times more effective with the proposed pre-training tasks. The extracted features from the OBF can also be used directly to classify children with and without autism
In the future, data analysts can use the pre-trained OBF directly in their cognition, psychology, marketing, and other studies that involve eye-tracking technologies. 

\begin{acks}
This study benefited from
perspectives and infrastructure provided by:
\grantsponsor{1}{NIH}{https://www.nih.gov/}
\grantnum[]{1}{K01 MH104739},
\grantnum[]{1}{R21 MH102572}, and 
\grantnum[]{1}{R03 MH092618};
\grantsponsor{2}{NSF Expedition in Socially Assistive Robotics}{https://www.nsf.gov/}
\grantnum[]{2}{\#113907};
\grantsponsor{3}{Simons Foundation}{https://www.simonsfoundation.org/}
\grantnum[]{3}{\#383661};
\grantsponsor{4}{Riksbankens Jubileumsfond}{https://www.rj.se}
\grantnum[]{4}{NHS14-1802:1}.

\end{acks}

\bibliographystyle{ACM-Reference-Format}
\bibliography{main}

\clearpage

\onecolumn

\section*{Appendix}

\subsection{Discussion on Data Pre-Processing}
The goal of data pre-processing is to unify data formats across different datasets rather than to improve the classification accuracy. The proposed data pre-processing contains two steps: (1) a simple linear transformation to unified units (visual degrees); (2) re-sampling to 60 Hz. 

The first step would not influence the model performance, as a neural network model can learn this transformation with one neuron. The 60 Hz frame rate is widely used in modern eye-tracking studies, as discussed in [23, 25]. Fixations are usually in the hundreds of milliseconds range, and 60 Hz data already provides enough information to identify fixations and saccades. We calculated the median Euclidean distance between 500 Hz and 60 Hz data and found that this “information loss” was less than 0.8 visual degrees -- roughly on the same scale as general experiment calibration error.

We acknowledge limitations of the re-sampling step, and future studies can explore better strategies to utilize information from higher frequency eye-tracking signals.

\begin{figure}[h!]
    \includegraphics[width=0.5\textwidth]{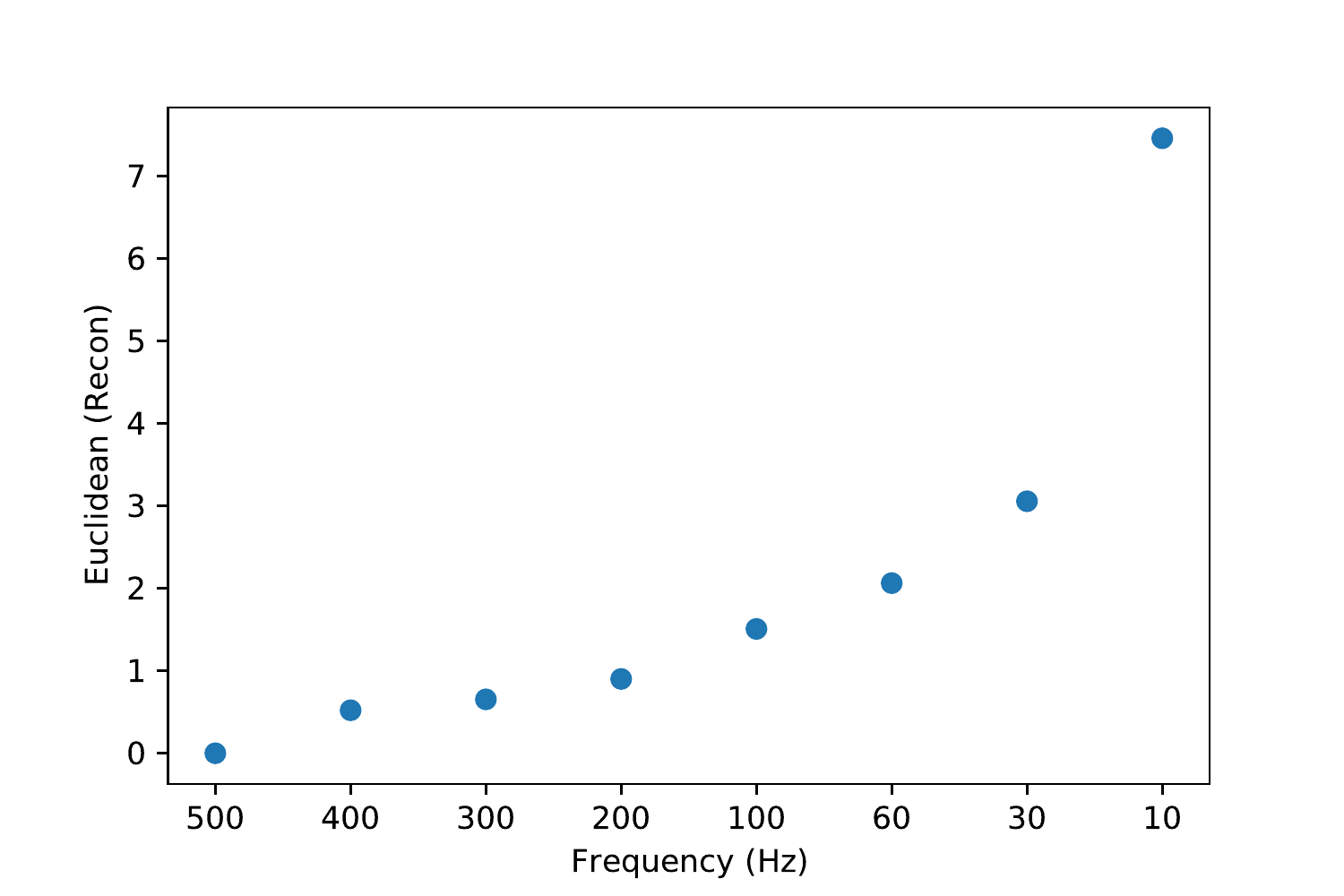}
\caption{Visualization on reconstruction error versus down-sample frequency for the pre-processing step. The x-axis is the downsampling frequency, and y-axis is the Euclidean reconstruction error (in visual degrees).}
\end{figure}

\newpage
\subsection{Downstream Application Details}
\begin{figure}[h!]
   \centering
   \begin{subfigure}[b]{0.50\textwidth}
        \includegraphics[width=0.9\textwidth]{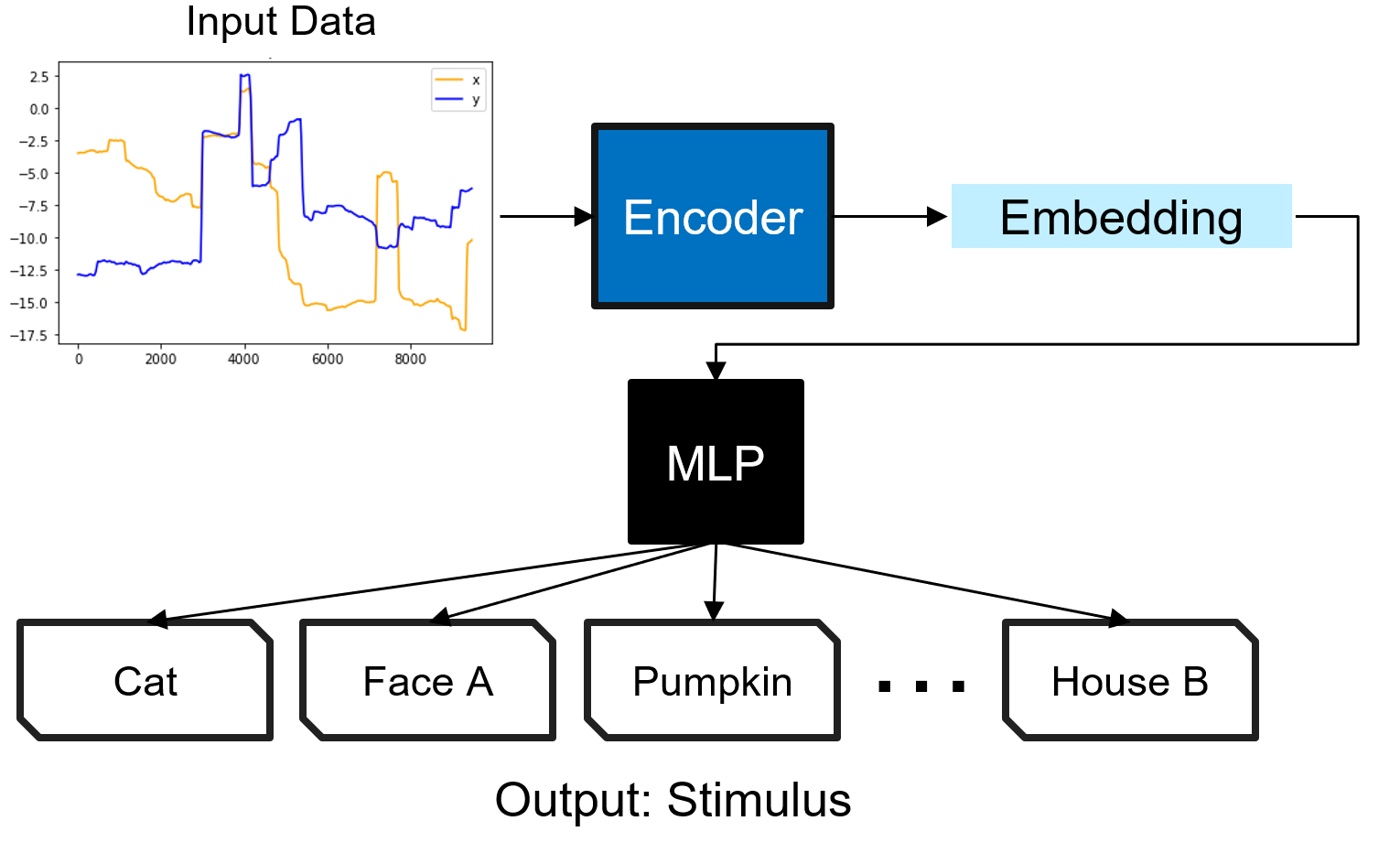}
        \caption{ Model Structure for MIT-1003 Supervised Learning Experiment }
        \label{fig:mit_model}
   \end{subfigure}
   \begin{subfigure}[b]{0.40\textwidth}
        \centering
        \includegraphics[width=0.9\textwidth]{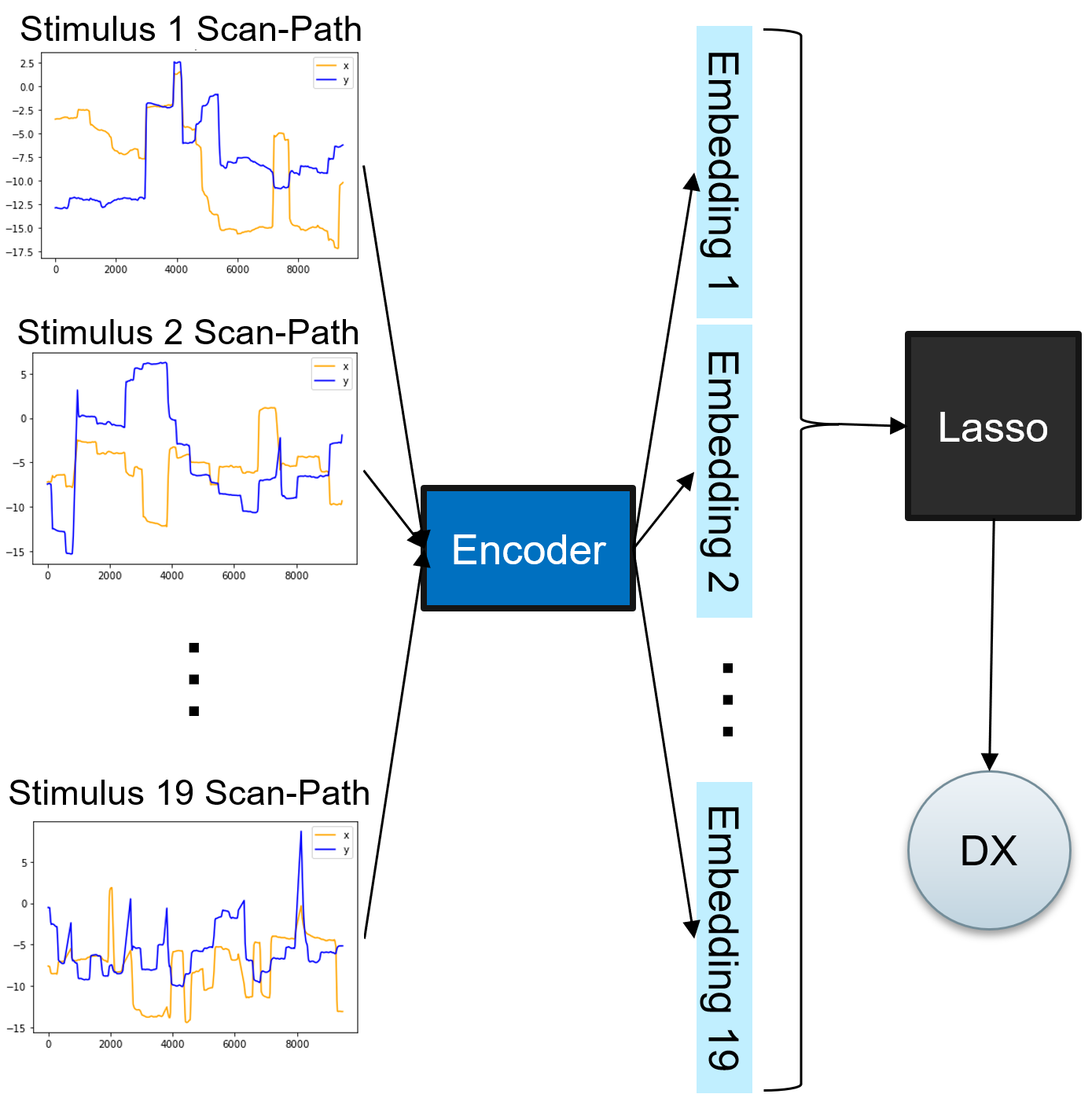}
        \caption{ Utilize the OBF encoder with lasso in autism classification}
        \label{fig:lasso_model}
   \end{subfigure}
   \caption{Model structures used in downstream applications.}
\end{figure}

\begin{figure*}[h!]
   \centering
   \includegraphics[width=0.7\textwidth]{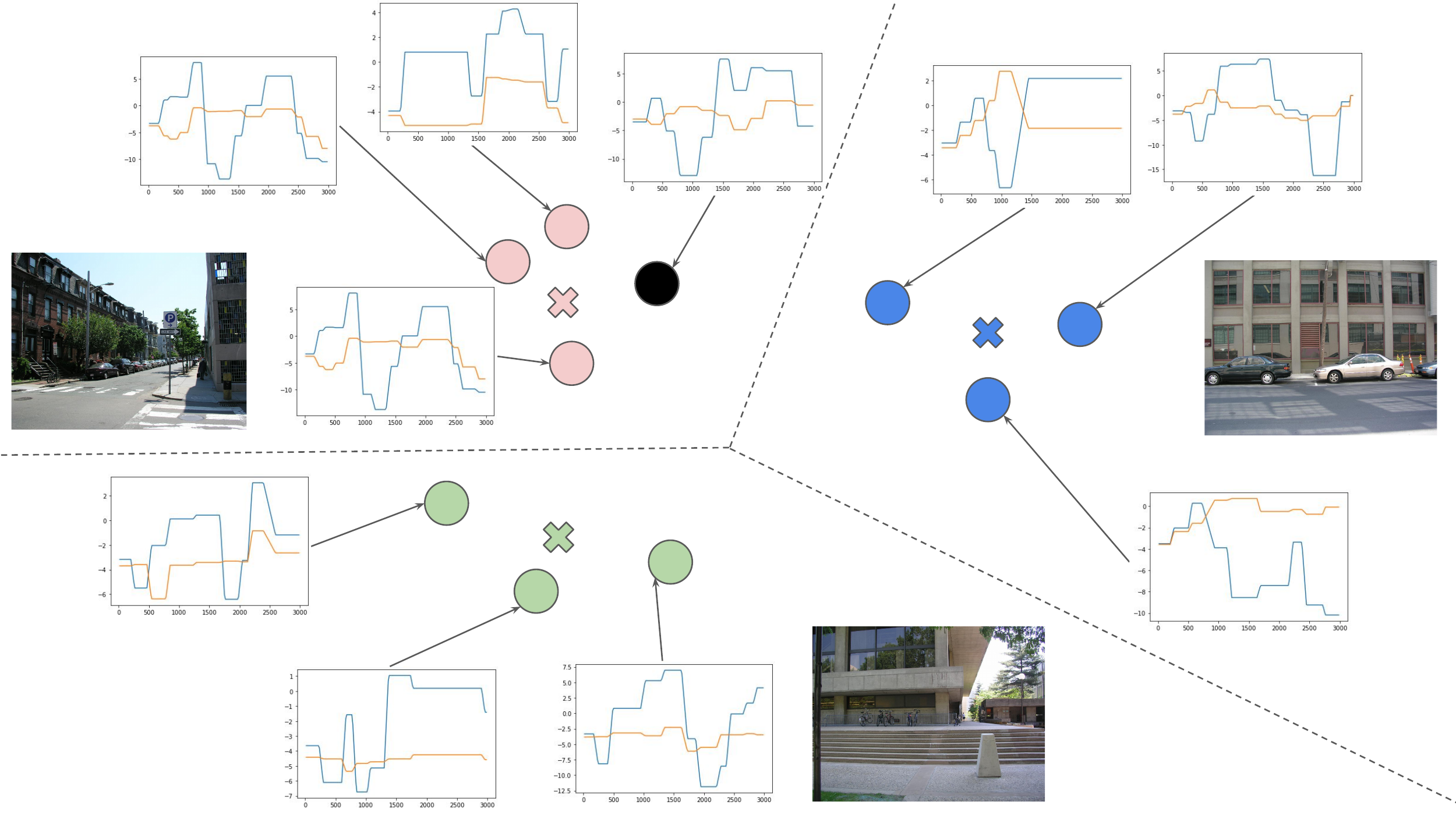}
   \caption{
   Illustration for the Meta-Learning Experiment for Stimulus Prediction: a 3-shot 3-way classification experiment is visualized here, where the goal is to predict which stimulus (i.e., one of the three street photos) the user was watching. We use red, green, and blue colors to represent these three classes (stimuli). Three scanpaths for each class are fed into our model, where the calculated embeddings are shown as circles for visualization purposes. The embedding center for each class is marked as a cross sign. During testing time, an unseen scanpath is fed into the network, and our model calculates its embedding (shown as the black circle); this scanpath is classified into the "red" class, because its embedding is closer to the center (i.e., red cross) of the red examples.
   }
   \label{fig:mit_proto_exp}
\end{figure*}

Figure \ref{fig:mit_model} and Figure \ref{fig:lasso_model} showed the model for supervised learning in Section \ref{sec:application}.
Figure \ref{fig:mit_proto_exp} showed the training and inference strategies in ProtoNet, a metric-based Meta-learning algorithm.

\newpage

\newpage
\subsection{Effects from Noise and Eye-tracking Experiment Errors}

To further illustrate the robustness of the proposed method, we use affine transformations to simulate real-world eye-tracking data noise (e.g., movement after calibration), including
\begin{enumerate}
    \item Add random offset to the signal (simulate: the screen/eye-tracker is slightly offset)
    \item Add random scale to the signal (simulate: the user is too close/far from the monitor)
    \item Add random rotation to the signal (simulate: the user's head is not parallel with the ground)
    \item Add random shear to the signal (simulate: the user's head is not parallel with the monitor)
    \item Add random noise to some points (simulate: the eye-tracker is not precise)
\end{enumerate}

If we include these errors in the data augmentation stage, then these noise and eye-tracking calibration errors will not affect the testing performance.

\subsection{Discussion on Pre-Training Tasks}
The goal of these pre-training tasks is to allow OBF to learn oculomotor behaviours from diverse unsupervised and supervised learning methods. The short embedded vector, calculated from the OBF encoder, can be easily stored and analyzed for various downstream applications. 
However, other model structures can be useful if a pre-training task is the ultimate goal for a given study. For instance,  traditional recurrent neural networks (rather than the seq-to-seq model) could be a better method if fixation identification is the ultimate goal, because the network does not need to store all temporal information in a short vector.

\begin{figure}[h!]
    \includegraphics[width=0.9\textwidth]{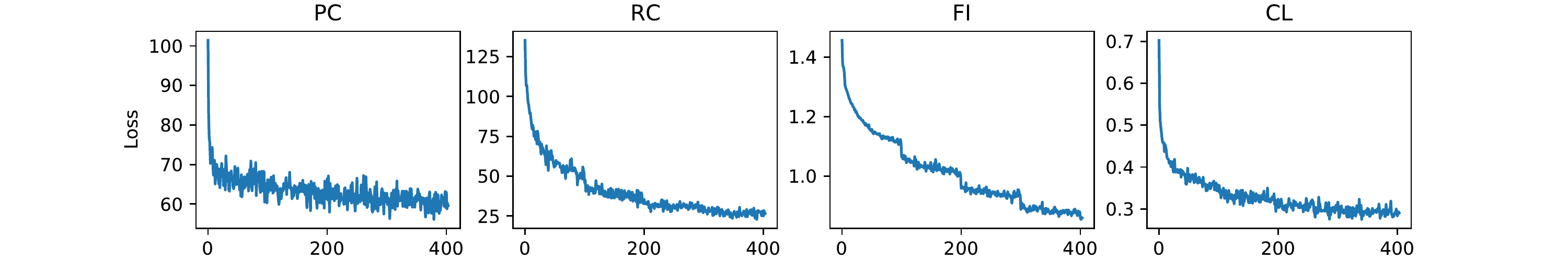}
\caption{All four pre-training tasks converged after 100 - 200 epochs. The x-axis is the epoch, and the y-axis is loss. Note that these pre-training tasks use different loss functions and hence the losses are not in the same scale. The predictive coding (PC) and reconstruction (RC) tasks use mean squared error of Euclidean distance, the fixation identification (FI) task uses weighted cross entropy, and the contrastive learning (CL) task use cross entropy loss.}
\label{fig:convergence}
\end{figure}

While multi-task learning (MTL) becomes a popular method to pre-train deep models, 
task selection becomes one of the most important steps. As discussed in Section \ref{sec:obe} and Section \ref{sec:pre_task_ablation}, we selected pre-training tasks both by human experts’ guidance and rigorous machine learning experiments. OBF converges well on all four pre-training tasks (as shown in Figure \ref{fig:convergence}), and ablation experiments (in Section \ref{sec:ablation}) show it also performs well on different downstream applications.  

\subsection{Results for Diverse Pre-Training Dataset}
As discussed in Section \ref{sec:diverse_datasets}, adding more datasets to the pre-training stage could create better OBF model. We show the ablation experiment results below in Table \ref{tab:predata}.

\begin{table}[h]
\centering
\begin{tabular}{c|ccc|ccc}
\toprule
\multirow{2}{*}{} & \multicolumn{3}{c|}{10-Shot Accuracy}          & \multicolumn{3}{c}{Autism-Classification}    \\
                  & 10-w        & 100-w       & 1003-w      & Acc          & F-1            & AUC           \\
\midrule
All Datasets & \textbf{0.65} & \textbf{0.48} & \textbf{0.31} & \textbf{0.80} & \textbf{0.88} & 0.83 \\
No MSU                & 0.53      & \textbf{0.48} & 0.29      & \textbf{0.80} & 0.87          & 0.78          \\
No C\&G-1             & 0.50          & 0.44     & 0.29           & 0.73         & 0.84          & 0.76          \\
No C\&G-2             & 0.57          & 0.44     & 0.28           & 0.78         & 0.86          & \textbf{0.84}           \\
\bottomrule
\end{tabular}
\caption{Ablation for pre-train datasets. 
With all pre-training datasets, the stimuli classification experiment showed the best performance.
For the 10-shot supervised learning experiment with MIT-1003 dataset, we evaluate the performance in 10-, 100-, 1003-way classification tasks.
For the Autism classification experiment, we evaluate the accuracy, F-1, and AUC scores.
}
\label{tab:predata}
\end{table}

\newpage
\subsection{Visualization for Contrastive Learning Task}
\begin{figure}[h!]
    \includegraphics[width=0.5\textwidth]{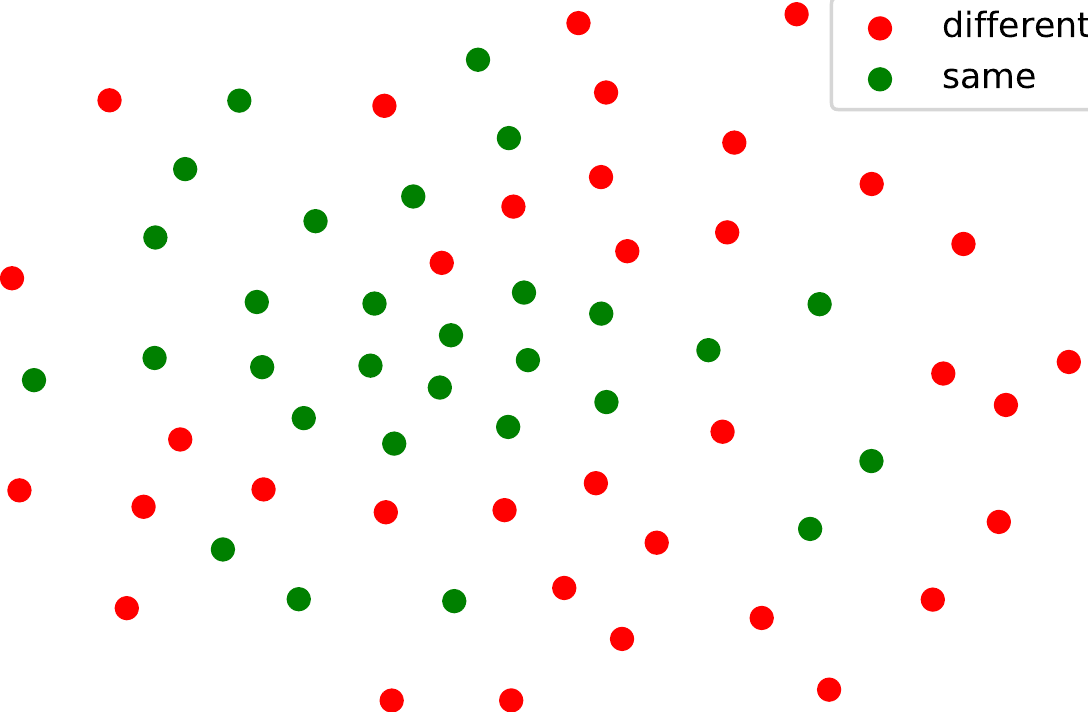}
\caption{We use t-distributed stochastic neighbor embedding (T-SNE) to reduce the dimension of embedding vectors for visualizing the contrastive learning (CL) results, where each dot in the figure corresponds to a vector $\mathcal{E}(x_1) - \mathcal{E}(x_2)$. When $x_1$ and $x_2$ are from the same scanpath, the vector is closer to the center (colored in green). }
\label{fig:tsne}
\end{figure}

For the contrastive learning task, We apply T-SNE on the embedding calculated from OBF and compared embeddings from two scanpaths in Figure \ref{fig:tsne}.

\end{document}